\documentclass[10pt,letterpaper]{article}

\usepackage{cogsci}

\cogscifinalcopy 

\usepackage[
  style=apa,
  natbib=true,
  annotation=false,
]{biblatex}
\usepackage{float} 

 \usepackage{graphicx}
 \usepackage{subcaption}
\usepackage{booktabs}
\title{Can Vision Replace Text in Working Memory?\\ Evidence from Spatial n-Back in Vision-Language Models}

\renewcommand\thefootnote{\fnsymbol{footnote}}
\author[1]{Sichu Liang\thanks{Equal Contribution. $^{\dag}$Corresponding author.}}
\author[2]{Hongyu Zhu\protect\footnotemark[1]}
\author[3]{Wenwen Wang}
\author[1]{Deyu Zhou\protect\footnotemark[2]}
\affil[1]{Southeast University}
\affil[2]{Shanghai Jiao Tong University}
\affil[3]{Carnegie Mellon University}

\begin{document}

\maketitle

\begin{abstract}

Working memory is a central component of intelligent behavior, providing a dynamic workspace for maintaining and updating task-relevant information. Recent work has used n-back tasks to probe working-memory-like behavior in large language models, but it is unclear whether the same probe elicits comparable computations when information is carried in a visual rather than textual code in vision-language models. We evaluate Qwen2.5 and Qwen2.5-VL on a controlled spatial n-back task presented as matched text-rendered or image-rendered grids. Across conditions, models show reliably higher accuracy and d' with text than with vision. To interpret these differences at the process level, we use trial-wise log-probability evidence and find that nominal 2/3-back often fails to reflect the instructed lag and instead aligns with a recency-locked comparison. We further show that grid size alters recent-repeat structure in the stimulus stream, thereby changing interference and error patterns. These results motivate computation-sensitive interpretations of multimodal working memory.

\end{abstract}

\setcounter{footnote}{0}
\renewcommand\thefootnote{\arabic{footnote}}

\section{Introduction}
Modern assistants and agents built on large language models (LLMs) and vision-language models (VLMs)~\citep{10.1561/0600000110,wang2024survey} often need to maintain and revise task state under distracting or stale context~\citep{liu2024lost,zhou2025towards}.
Driven by context-length and memory constraints, an emerging practice is to treat visual representations as a compact ``memory substrate'': systems render task state or interaction history into images and rely on a VLM to read and use this visualized memory \citep{xing2025visioncentric,wei2025deepseek,feng2026agentocr}.
This practice implicitly assumes that replacing text tokens with visual tokens as a memory carrier does not change the model’s update-and-compare strategy.

Such state maintenance and revision is commonly framed as working memory (WM), the limited set of task-relevant representations kept accessible during ongoing behavior~\citep{BADDELEY197447,baddeley2000episodic}.
In cognitive science, WM is not only short-term retention but coordinated control processes---updating, attentional selection, and resistance to interference---supporting goal-directed behavior and reasoning \citep{baddeley2020working,kane2000working,jonides2006brain}.
When we apply WM probes to LLMs and VLMs, a key question is \emph{what level of computation the probe is actually measuring}: do apparent ``WM limits'' reflect a failure of online updating, or a shift in strategy induced by how information is represented?



A widely used probe for WM is the \textit{n}-back task \citep{kirchner1958age}, in which a participant observes a stimulus stream and reports whether the current item matches the one presented $n$ steps earlier.
Spatial \textit{n}-back uses locations on a grid and has been widely used in behavioral and neuroimaging work \citep{owen2005n,JANSMA2000688,mcmillan2007self}.
Although larger $n$ is intended to increase demands on online updating and temporal context binding, decades of human research emphasize that \textit{n}-back performance reflects multiple components---including interference control and strategy choice---rather than a single capacity factor \citep{harbison2012n, frost2021n, postle2005effects,jaeggi2008improving}.

\begin{figure}
    \centering
    \includegraphics[width=0.98\linewidth]{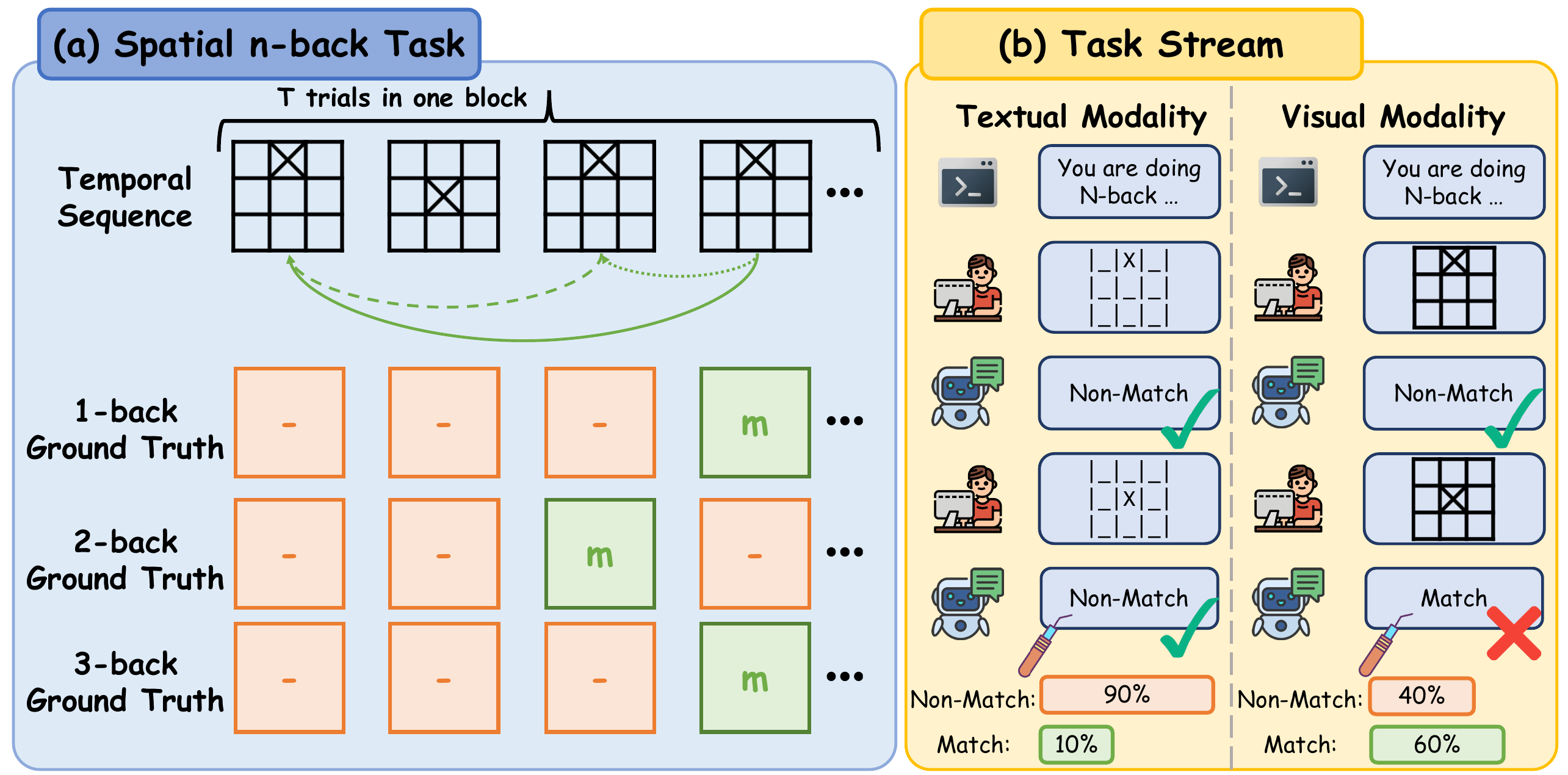}
    \caption{\textbf{Spatial \textit{n}-back task and multimodal presentation.}
    (a) A spatial \textit{n}-back block consists of a temporal sequence of grids; the target cell is marked by \texttt{X}. A trial is labeled as \texttt{Match} if the current location matches the location \(n\) steps back.
    (b) The same trial stream is presented in two formats: a text-rendered grid (text modality) and an image-rendered grid (vision modality), with trial-wise decision evidence extracted from the forced-choice responses.}
    \label{fig:small}
\end{figure}

Crucially, varying the \emph{representational code} can change which computations are supported or preferred; in Marr’s terms, holding the task objective fixed while changing the code can expose shifts at the mic level \citep{marr2010vision, brady2010encoding}.
Consistent with this, human \textit{n}-back performance depends on stimulus modality and presentation format \citep{amon2018auditory}.
Thus, the same nominal \textit{n}-back rule may be realized by different strategies (e.g., lagged binding vs.\ recency-biased comparisons), especially under higher load or interference.
This raises a concrete question: when the task objective is held constant, does changing the representational code (text vs.\ vision) preserve the computations that spatial \textit{n}-back is intended to elicit?
We therefore ask (i) whether representational code yields a systematic performance difference and (ii) whether process diagnostics indicate shifts in temporal-context binding and susceptibility to proactive interference from recent history~\citep{jonides2006brain, destefano2020influences}.

Recent work has begun to apply classic WM paradigms to large language models (LLMs), using \textit{n}-back tasks to quantify how performance changes with nominal load and how strongly it depends on task formulation \citep{gong2024working,zhang2024working}.
A recurring challenge is interpretability: poor performance at higher \textit{n} may reflect not only memory limitations but also failures to maintain the intended task set, including drift toward simpler comparison strategies \citep{hu-lewis-2025-language}.
Related work further suggests that task formulation and input complexity can interact with nominal load, complicating capacity-style interpretations \citep{hong-etal-2025-exploring}.
However, this emerging literature has largely operationalized WM with text-based inputs, leaving open whether the same probes elicit comparable \emph{intended computations} when the representational code itself is changed.

Modern vision-language models (VLMs) offer a more controlled testbed for representational-code effects, since they can process both text and images within a unified architecture \citep{hurst2024gpt,team2024gemini}, and open-weight systems enable controlled and reproducible multimodal evaluation \citep{liu2023visual,10.1145/3746027.3755002}.
Because VLMs typically convert both images and text into token sequences and then rely on largely shared language-model components \citep{qwen2.5-VL}, changing the representational code may change how online control processes (updating, temporal binding, interference resistance) are implemented in WM probes.


Here we examine this question using a controlled spatial \textit{n}-back paradigm implemented in two matched formats: a text-rendered grid and an image-rendered grid (Figure~\ref{fig:small}).
We compare performance across representational codes (text vs.\ vision) and observe systematic differences that motivate process-level diagnostics:
(i) a \emph{lag-scan} analysis that tests whether trial-wise decision evidence aligns with the instructed comparison lag versus a recency-locked alternative; and
(ii) a \emph{grid-size} manipulation (stimulus set size) that changes the rate of recent-repeat lures, allowing us to characterize proactive interference from recent history.

Together, these results provide a multimodal characterization of WM-like behavior in LLMs and VLMs.
Empirically, we show that replacing text with vision in a matched spatial \textit{n}-back probe can lead to substantial performance differences.
Conceptually, we highlight that representational code and stimulus statistics can change the computation realized by a WM probe: multimodal evaluation should test not only end-point performance, but also whether models implement the temporal-context binding and interference-sensitive control processes that such paradigms are intended to elicit.

\section{Method}

\subsection{Models as Participants}
We treat models as experimental participants and evaluate their behavior under controlled task conditions.
Our main experiments use the text-only model \textsc{Qwen2.5-7B-Instruct}~\citep{qwen2.5} and its vision-language counterpart \textsc{Qwen2.5-VL-7B-Instruct}~\citep{qwen2.5-VL}. Throughout the paper, we refer to models without the ``-Instruct'' suffix for brevity.
Unless otherwise specified, we use deterministic decoding (temperature $=0$) to eliminate sampling variability and enable trial-by-trial reproducibility.

\subsection{Task, Stimuli, and Design}
We use a spatial \textit{n}-back task defined over a discrete $N{\times}N$ grid.
Each trial $t$ presents a single target location (one occupied cell), and the model judges whether this location matches the one shown $n$ trials earlier ($t-n$). 
Responses are binary (\textit{match} vs.\ \textit{non-match}).
We vary memory load $n\in\{1,2,3\}$ and grid size $N\in\{3,4,5,7\}$.
For the first $n$ trials in each block ($t \le n$), the $n$-back comparison is undefined; by convention we treat these warm-up trials as \textit{non-match}, as explicitly stated in the instruction prompt for models.

\paragraph{Text-grid condition.}
As shown in Figure~\ref{fig:small}, the $N{\times}N$ grid is rendered as an ASCII layout in the prompt.
The target location is marked with \texttt{X} and all other cells are filled with \texttt{\_}, with $N$ rows separated by line breaks.
This format is used for both text-only LLMs and VLMs.

\paragraph{Vision-grid condition.}
The same $N{\times}N$ grid is rendered as an image showing a lattice with a single \texttt{X} marker in the target cell.
The image is the only trial-specific stimulus; surrounding textual instructions are identical to the text-grid condition.
Images are generated programmatically to control layout and avoid incidental cues, using a fixed $256{\times}256$ canvas with consistent line thickness and font size across trials. \footnote{The text-grid condition contains 26/38/52/86 text tokens for grid sizes 3/4/5/7, respectively. In contrast, the vision-grid condition yields a constant 86 visual tokens across all grid sizes.}

\paragraph{Stimulus readability check (location recognition).}
To verify that the grid stimulus is legible under each input format, we run a single-trial location recognition probe.
The model is shown one grid and reports the target cell as a (row, column) coordinate; we compute exact-match accuracy over 200 randomly generated grids for each size and format.
As shown in Table~\ref{tab:loc_recog_acc}, recognition remains high in the vision-grid condition, suggesting that the main modality gap is unlikely to be driven by stimulus unreadability.

\paragraph{Trial sequence generation.}
For each $(N,n)$ condition, we pre-generate trial sequences and organize them into blocks.
A \emph{trial} is one stimulus--response event, and a \emph{block} is a contiguous sequence of trials presented without resetting the context.
Within each block, match and non-match trials occur in a fixed $1{:}1$ ratio; each condition comprises $B{=}50$ independently generated blocks of $T{=}24$ trials (1{,}200 trials total).
To isolate the effect of representational code, the text-grid and vision-grid conditions share the same pre-generated sequences, and prompts are identical except for the stimulus format. \footnote{Trial sequences are fixed for reproducibility; anonymized code is available at \url{https://anonymous.4open.science/r/vision_wm-EFB0}.}

\begin{table}[t]
    \centering
    \resizebox{\linewidth}{!}{
    \begin{tabular}{lcccc}
        \toprule
        \multicolumn{1}{l}{Accuracy (\%)} &
        \multicolumn{1}{c}{$3{\times}3$} &
        \multicolumn{1}{c}{$4{\times}4$} &
        \multicolumn{1}{c}{$5{\times}5$} &
        \multicolumn{1}{c}{$7{\times}7$} \\
        \midrule
        Qwen2.5-7B-Instruct (text)            & 80.50 & 73.00 & 62.00 & 55.00 \\
        Qwen2.5-VL-7B-Instruct (text)         & 70.00 & 63.50 & 58.00 & 53.50 \\
        Qwen2.5-VL-7B-Instruct (vision)       & 78.50 & 82.00 & 77.00 & 80.50 \\
        \bottomrule
    \end{tabular}
    }
    \caption{Location recognition accuracy (200 random grids per size and format). Used as a stimulus-readability check.}
    \label{tab:loc_recog_acc}
\end{table}

\subsection{Procedure}
The task is administered sequentially in a trial-by-trial format analogous to human \textit{n}-back procedures: within each block, stimuli are presented in order and the model responds on every trial.
Each block begins with a brief instruction and a forced-choice response rule: output $\mathtt{m}$ for \texttt{Match} and $\mathtt{-}$ for \texttt{Non-match}.
On each trial, the model is shown the current stimulus (text grid or image) and is constrained to generate exactly one of the two labels, $\{\mathtt{m},\mathtt{-}\}$, with no additional text.

For each trial context $x_t$, we record (i) the predicted label and (ii) token-level log-probabilities for both candidate labels at the forced-choice output position, which we later convert into a graded match-evidence score (Eq.~\ref{eq:evidence_score}).

\subsection{Dependent Measures}

\paragraph{Behavioral measures (Accuracy, $H/FA$, and $d'$).}
We report standard \textit{n}-back measures that are routinely used in working-memory studies \citep{hautus2021detection, owen2005n}.
From the forced-choice labels, we compute accuracy (proportion correct), the hit rate $H$ (proportion of ground-truth match trials labeled $\mathtt{m}$), and the false-alarm rate $FA$ (proportion of ground-truth non-match trials labeled $\mathtt{m}$).
Sensitivity is summarized by $d'$, a standard signal-detection index computed from $H$ and $FA$ (with the loglinear correction), where larger $d'$ indicates better discriminability.

\paragraph{Evidence-based discriminability (AUC).}
For open-weight models, token log-probabilities allow us to treat label preference as a continuous decision variable, a common practice in likelihood-based human and LLM evaluation \citep{brown2020language, kiani2009representation, o2022measuring}.
We define match evidence score for trial $t$ as the log-likelihood difference
\begin{equation}
    s_t \triangleq
    \underbrace{\log p_\theta\!\left(y=\mathtt{m}\mid x_t\right)}_{\text{match evidence}}
    \;-\;
    \underbrace{\log p_\theta\!\left(y=\mathtt{-}\mid x_t\right)}_{\text{non-match evidence}} ,
    \label{eq:evidence_score}
\end{equation}
where probabilities are taken at the forced-choice output position.
We compute the area under the ROC curve (AUC) using $s_t$ as the decision variable \citep{fawcett2006introduction, yonelinas1994receiver}. AUC summarizes how well $s_t$ separates match from non-match trials across  criterion (decision thresholds), providing a threshold-free measure of discriminability.

\begin{figure*}[t]
    \centering
    \includegraphics[width=\linewidth]{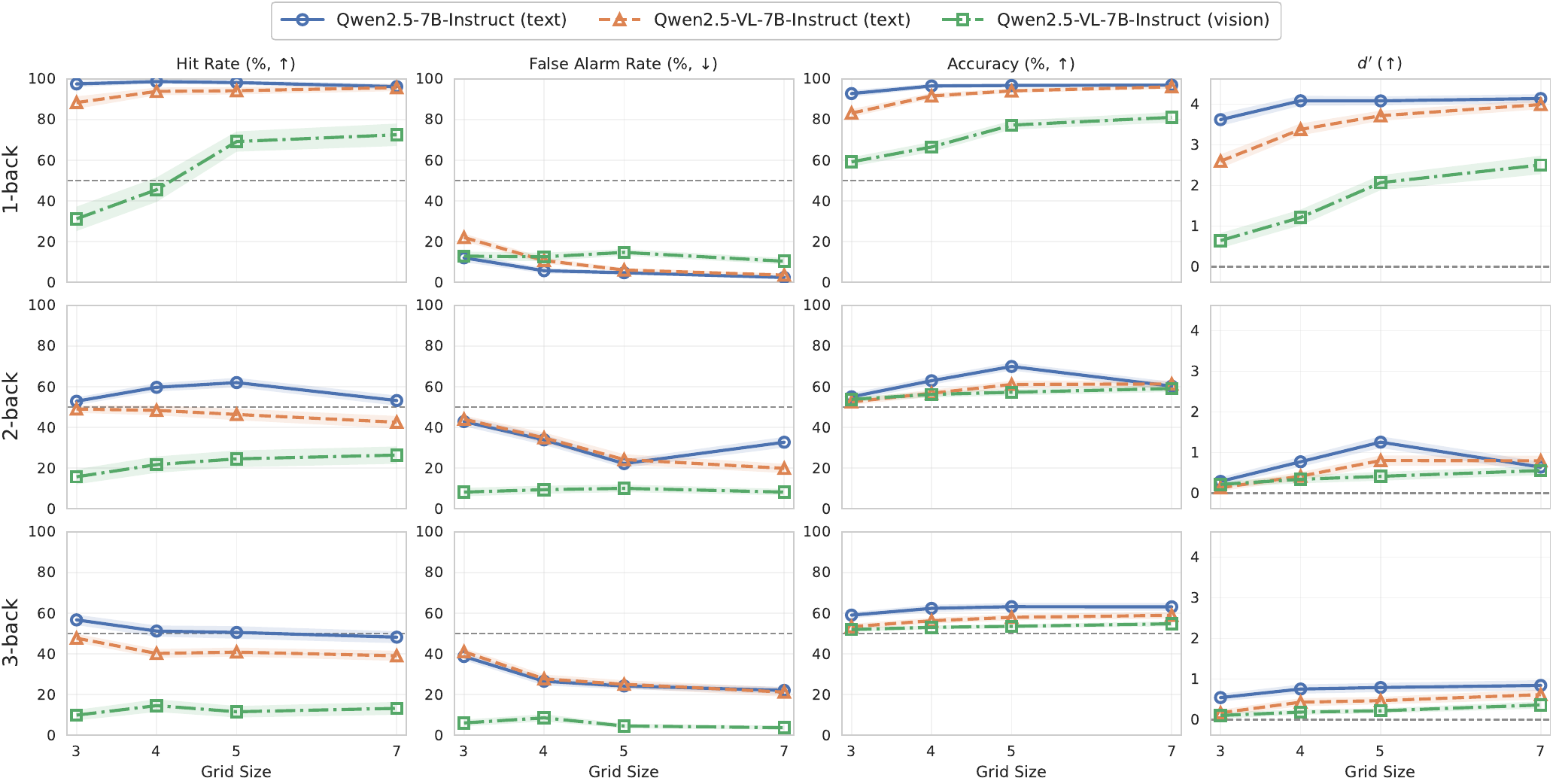}
    \caption{\textbf{Main results on spatial \textit{n}-back.}
Hit rate, false-alarm rate, accuracy, and $d'$ versus grid size $N$ for $n{=}1,2,3$ (rows).
Curves compare \textsc{Qwen2.5-7B} (text-grid), \textsc{Qwen2.5-VL-7B} (text-grid), and \textsc{Qwen2.5-VL-7B} (vision-grid); bands show $\pm1$ SEM (standard error of the mean) across blocks.
Dashed lines mark chance-level references (50\% for rates/accuracy; $d'{=}0$).}
    \label{fig:main}
\end{figure*}

\begin{figure}[t]
    \centering
    \includegraphics[width=\linewidth]{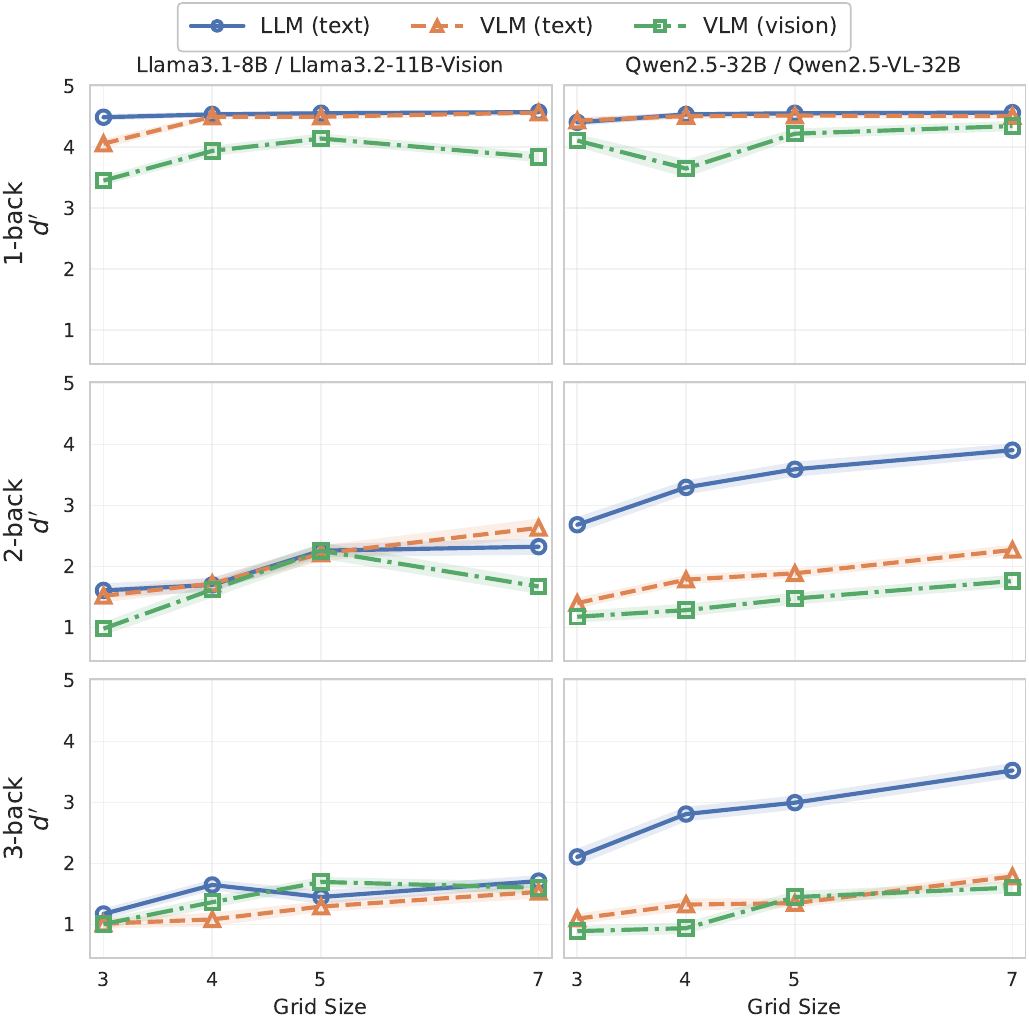}
    \caption{\textbf{Robustness across model families and scale.}
Sensitivity ($d'$) versus grid size $N$ for $n{=}1,2,3$ (rows) in two LLM/VLM pairs (left: \textsc{Llama3.1-8B}/\textsc{Llama3.2-11B-Vision}; right: \textsc{Qwen2.5-32B}/\textsc{Qwen2.5-VL-32B}).
For each pair, we compare LLM(text-grid), VLM(text-grid), and VLM(vision-grid); bands show $\pm1$ SEM across blocks.}

    \label{fig:size}
\end{figure}

\section{Results}
\subsection{Main Results}

Figure~\ref{fig:main} summarizes spatial \textit{n}-back performance for three evaluation conditions: a text-only LLM (\textsc{Qwen2.5-7B}, text-grid), the corresponding VLM given the same text-rendered grids (\textsc{Qwen2.5-VL-7B}, text-grid), and the same VLM given rendered images of the grids (\textsc{Qwen2.5-VL-7B}, vision-grid). We report standard measures (hit rate, false-alarm rate, accuracy, and sensitivity $d'$) across grid sizes $N\in\{3,4,5,7\}$ (x-axis) and memory loads $n\in\{1,2,3\}$ (rows).

\paragraph{Modality effect.}
Across loads and grid sizes, performance is consistently highest when the task stream is presented as text to the text-only model, slightly lower for the VLM under the same text-grid format, and substantially lower when the VLM receives the vision-grid format. This modality gap is most apparent in hit rates: in the vision-grid condition, models miss many ground-truth match trials even at 1-back, and hit rates remain far below the text-grid conditions at higher loads. False-alarm rates show a different pattern: under 2- and 3-back, text-grid conditions exhibit non-trivial false alarms, whereas the vision-grid condition remains comparatively conservative (lower false alarms) but at the cost of markedly reduced hits, yielding low overall sensitivity $d'$.

\paragraph{Load effect.}
Increasing nominal load produces a sharp reduction in discriminability.
Under 1-back, both text-grid conditions achieve high accuracy and large $d'$, while the vision-grid condition remains substantially lower.
Under 2-back and 3-back, sensitivity drops toward near-chance levels in multiple settings, with the vision-grid condition showing particularly poor discriminability (where $d'$ stays close to zero over grid sizes).
Critically, this near-chance sensitivity is driven less by elevated false alarms than by a collapse in hit rates: the model becomes increasingly reluctant to endorse \texttt{Match}, yielding low $H$ alongside relatively low $FA$, and therefore weak separation between match and non-match despite only modest changes in accuracy.

\paragraph{Grid-size effect.}
Performance also varies with grid size. Across conditions, moving from smaller grids to larger grids tends to improve accuracy and $d'$, counterintuitively for a larger state space. Notably, the vision-grid condition benefits from increased grid size but remains well below the text-grid baselines, motivating further analyses of how stimulus structure shapes interference and decision behavior.

\subsection{Robustness Across Model Families and Scale}

Figure~\ref{fig:size} tests whether the main qualitative patterns generalize beyond the 7B setting, using two additional LLM/VLM pairs: \textsc{Llama3.1-8B} with \textsc{Llama3.2-11B-Vision}~\citep{dubey2024llama} (left) and \textsc{Qwen2.5-32B} with \textsc{Qwen2.5-VL-32B} (right). We focus on sensitivity ($d'$) and evaluate each VLM under both text-grid and vision-grid inputs.

Across both model families, three trends replicate. First, sensitivity decreases substantially with nominal load $n$, consistent with increased demands on lagged temporal binding. Second, for a fixed model family and load, performance follows the same modality ordering as in the main results: text-grid inputs yield higher $d'$ than vision-grid inputs, with VLM(text) typically lying between the LLM(text) and VLM(vision). Third, increasing grid size generally improves $d'$---most clearly at higher loads---suggesting that stimulus structure can reduce effective interference even when nominal load is held constant.

Model family and scale primarily modulate the \emph{magnitude} of the modality gap rather than its direction. In the \textsc{Qwen2.5-32B} pair, the separation between text-grid and vision-grid tends to widen at larger grids (especially for $n{=}2,3$), whereas the \textsc{Llama} pair shows the same ordering with a smaller gap. Overall, the qualitative patterns are stable across families and scales, while the size of the text--vision gap varies.

\begin{figure}[t]
    \centering
    \includegraphics[width=\linewidth]{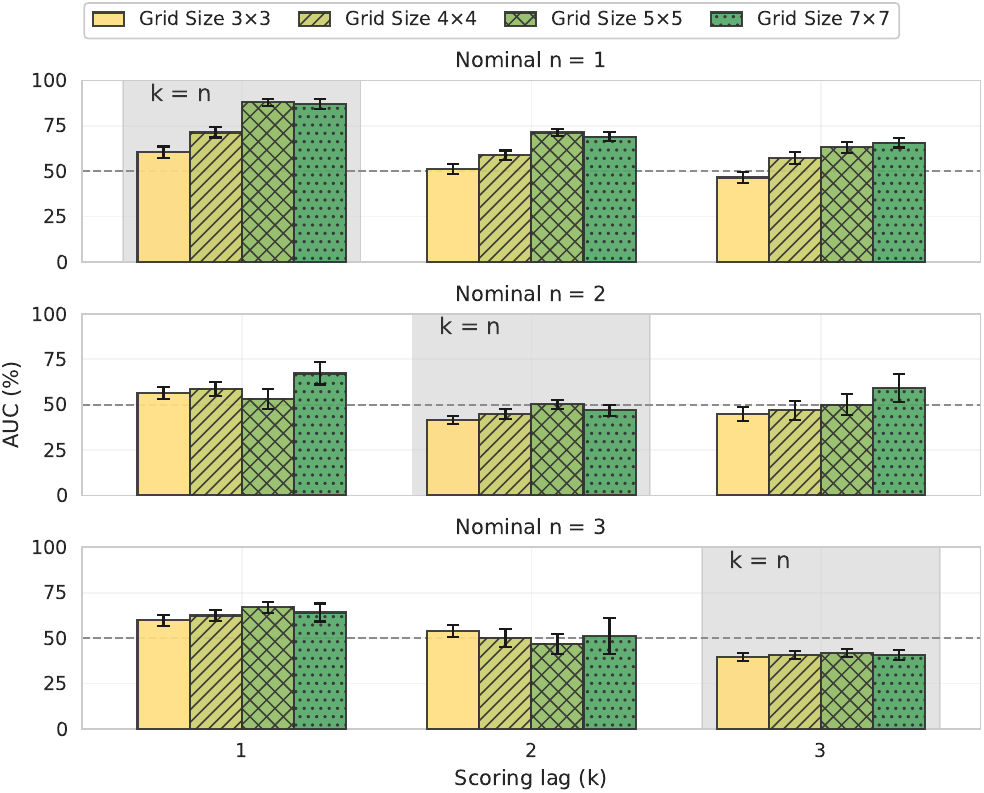}
    \caption{\textbf{Lag-scan diagnostic.} Using the evidence $s_t$, we compute AUC against match/non-match labels defined by assumed lags $k\in\{1,2,3\}$ for each nominal instruction $n$. AUC at the instructed definition ($k{=}n$, shaded) drops toward chance for $n{=}2,3$. Across $n$, AUC peaks at $k{=}1$ rather than $k{=}n$.}

    \label{fig:ab1}
\end{figure}

\subsection{Process Diagnostics I: Criterion Bias or Separability}
\label{sec:lag_scan}
The main results show poor discriminability at nominal memory load $n{=}2,3$, especially in the vision-grid condition.
To interpret these failures, we use the trial-wise match-evidence score $s_t$ (Eq.~\ref{eq:evidence_score}) and ask two process questions: (i) is performance limited mainly by a conservative decision criterion, or by weak evidence separability; and (ii) if evidence is weak under the instructed rule, which temporal comparison does it actually support?

\paragraph{Is poor performance mainly a conservative criterion?}

The vision-grid condition shows many missed matches (low $H$) and few false alarms (low $FA$; Fig.~\ref{fig:main}), consistent with a conservative, non-match-biased response policy.
To test whether matches and non-matches remain separable despite this bias, we use the match-evidence score $s_t$ (Eq.~\ref{eq:evidence_score}) and compute AUC, a threshold-free measure of discriminability.
We score evidence under a \emph{scoring lag} $k$: a trial is labeled \textit{match} if the current location repeats the location $k$ trials earlier, and \textit{non-match} otherwise.
The instructed task corresponds to $k{=}n$ (shaded in Fig.~\ref{fig:ab1}).
If failures were primarily due to an overly conservative threshold, AUC at $k{=}n$ would remain high; instead, AUC at $k{=}n$ drops toward chance at nominal $n{=}2,3$, indicating weak evidence separation that cannot be recovered by shifting the decision criterion, matching $d'\!\approx\!0$.

\paragraph{Lag-scan: evidence tracks 1-back rather than the instructed lag.}
We then keep the same evidence sequence $\{s_t\}$ and recompute AUC for $k\in\{1,2,3\}$, effectively asking which temporal comparison the model's graded evidence supports.
Across nominal $n\in\{1,2,3\}$, evidence--label alignment consistently peaks at $k{=}1$ rather than at the instructed lag $k{=}n$ (Fig.~\ref{fig:ab1}), suggesting systematic \emph{task-set drift} toward a recency-locked (1-back) comparison.

\begin{figure*}[t]
    \centering

    \begin{subfigure}[t]{0.24\linewidth}
        \centering
        \includegraphics[width=\linewidth]{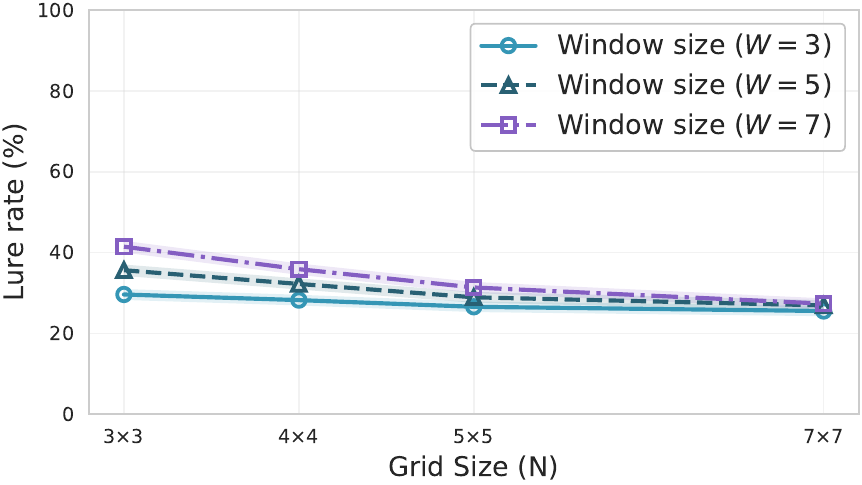}
        \caption{\textbf{Lure rate decreases with grid size acorss window size.}}
        \label{fig:fig4A}
    \end{subfigure}\hfill
    \begin{subfigure}[t]{0.24\linewidth}
        \centering
        \includegraphics[width=\linewidth]{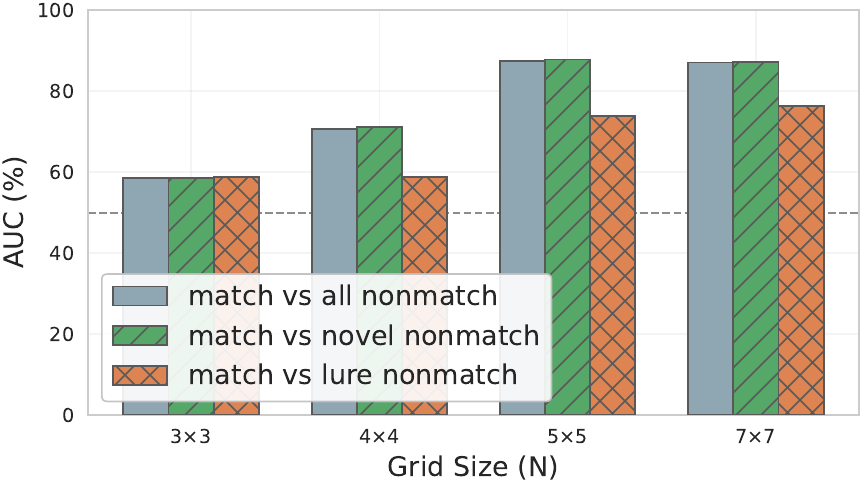}
        \caption{\textbf{Recent-repeat lures selectively reduce discriminability.}}
        \label{fig:fig4B}
    \end{subfigure}\hfill
    \begin{subfigure}[t]{0.24\linewidth}
        \centering
        \includegraphics[width=\linewidth]{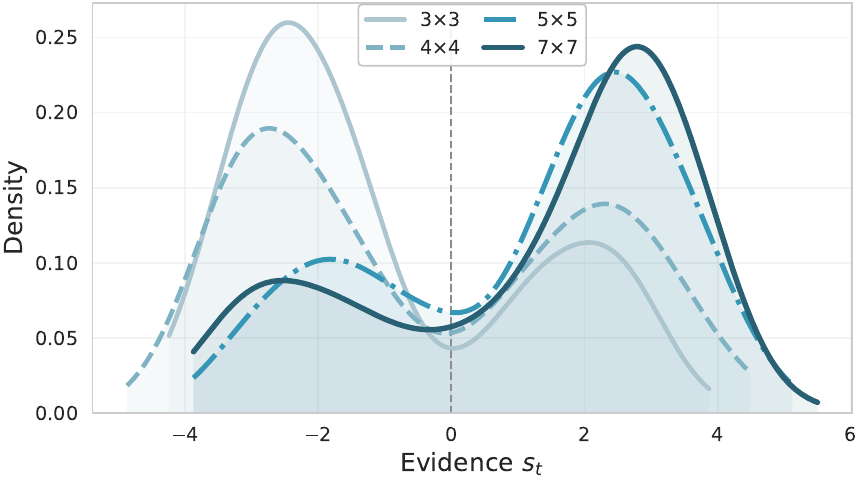}
        \caption{\textbf{Small grids induce a low-evidence mode on match trials.}}
        \label{fig:fig4C}
    \end{subfigure}\hfill
    \begin{subfigure}[t]{0.24\linewidth}
        \centering
        \includegraphics[width=\linewidth]{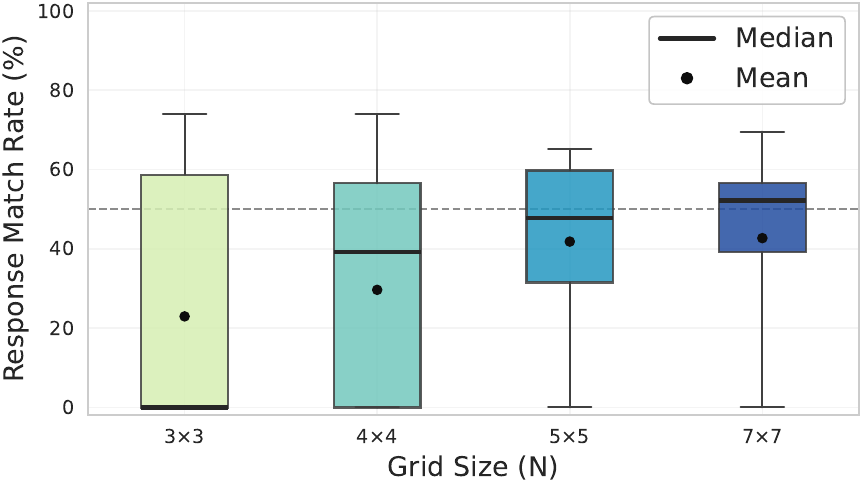}
        \caption{\textbf{Block-wise match response collapses in small grids.}}
        \label{fig:fig4D}
    \end{subfigure}

\caption{\textbf{Explaining the grid-size advantage via recent-repeat interference (in 1-back).}
(A) Recent-repeat lure rate decreases with grid size $N$ (in all $W\in\{3,5,7\}$).
(B) \emph{Lure non-matches}: discriminability (AUC from $s_t$) is lower when contrasting matches against lure non-matches than against novel non-matches for $N\ge4$, indicating that recent repeats make \emph{non-match} trials more confusable.
(C) \emph{Lure matches}: in small grids, match trials in a recent-repeat context receive lower (often negative) evidence $s_t$, indicating suppressed match evidence under dense repetition.
(D) Small grids show a block-level collapse toward always predicting non-match (match-response rate distribution; $3{\times}3$ median $=0$).
Bands/error bars show $\pm1$ SEM across blocks; dashed lines mark 50\% for AUC (\%) and match response rate.}
    \label{fig:ab2}
\end{figure*}

\subsection{Process Diagnostics II: Recent-Repeat Interference}
\label{sec:grid_effects}

We next examine why discriminability tends to improve with grid size in 1-back.
Because nominal 2-back and 3-back are often near chance (low $d'$ and AUC $\approx 50\%$), breaking errors into finer subtypes is less informative there.
We therefore focus on nominal 1-back and analyze how \emph{recent repeats within a block} shape decisions (Fig.~\ref{fig:ab2}).

Fix a recency window of length $W$ within each block. A trial is a \emph{recent-repeat lure} if the current location repeats any location in the previous $W$ trials \emph{excluding} the instructed comparison position $t-n$.
We use this definition to compute overall lure rates for multiple window sizes (Fig.~\ref{fig:fig4A}, $W\in\{3,5,7\}$); for subtype analyses we fix $W{=}3$ and distinguish \emph{lure non-matches} (ground-truth non-match trials that are lures) from \emph{lure matches} (ground-truth match trials that also repeat elsewhere within the window; Fig.~\ref{fig:fig4B}--\ref{fig:fig4C}).

\paragraph{Lures make \emph{non-match} trials more confusable (Fig.~\ref{fig:fig4B}).}
Using the evidence score $s_t$, we compute AUC while varying the negative set (all non-matches vs.\ \emph{novel} non-matches without recent repeats vs.\ lure non-matches; $W{=}3$).
For $N\ge4$, discriminability is consistently lower when negatives are lure non-matches (Fig.~\ref{fig:fig4B}), indicating that recent repeats selectively increase confusion on the \emph{non-match} side.
Together with the reduced lure frequency at larger $N$ (Fig.~\ref{fig:fig4A}), this explains a substantial portion of the grid-size advantage.

\paragraph{In small grids, repeats also suppress evidence on \emph{match} trials and can trigger block-level collapse (Fig.~\ref{fig:fig4C}--\ref{fig:fig4D}).}
Restricting to ground-truth match trials, we compare evidence distributions for lure matches trials among grid sizes (again $W{=}3$; Fig.~\ref{fig:fig4C}).
Small grids show a pronounced low-evidence mode on lure matches, indicating suppressed match evidence under dense repetition.
At the response level, this coincides with a striking block-wise collapse for $3{\times}3$: more than half of blocks predict non-match on essentially all trials (median match response rate $=0 \%$; Fig.~\ref{fig:fig4D}).

Taken together, these results suggest a two-sided mechanism for the grid-size effect.
In larger grids, lures are rarer and primarily act by making a subset of non-match trials harder.
In small grids, dense repetition affects both sides of the decision: lure non-matches remain confusable while lure matches lose evidence, and the combined pressure can drive a degenerate ``always non-match'' response policy at the block level.

\section{Discussion}
\label{sec:discussion}

This work asks a simple representational question with a classic probe: when the \emph{task objective} is held fixed in spatial \textit{n}-back, what changes when the same information is delivered as text versus vision?
Across model families and scales, we observe a reliable ordering in discriminability (LLM text-grid $>$ VLM text-grid $>$ VLM vision-grid), and discriminability drops sharply as nominal load increases from 1-back to 2/3-back.
Crucially, a lag-scan diagnostic indicates that under nominal 2/3-back, graded evidence is often best aligned with a 1-back relation, consistent with a recency-locked comparison policy rather than the instructed lagged binding \citep{schmiedek2009interference}.
Taken together, these patterns argue that representational code can shift the \emph{effective algorithm} a model implements in a working-memory probe, not merely the ease of reading the stimulus~\citep{marr2010vision}.

The same temporal-binding perspective clarifies the ``grid-size advantage'' in \textit{n}-back.
A key failure source here is interference from \emph{recent repeats} (lures) in the local temporal context \citep{kane2007working, jonides2006brain}.
Recent work reports this advantage in both humans and LLMs, and discusses it in terms of perceptual affordances in humans (e.g., eye-movement / visual-search ease in larger displays) and increased input length in text-rendered grids for LLMs \citep{zhang2024working}.
Our results support a complementary, mechanism-level account by \emph{dissociating} grid size from input length.

In the vision-grid condition, all grids are rendered on a fixed canva, yielding a constant visual token count under VLM tokenization across grid size $N$.
Nevertheless, discriminability improves with larger grids.
This dissociation rules out a trivial ``more tokens / longer prompts'' explanation and instead points to proactive interference: larger state spaces reduce accidental recurrences and therefore reduce confusable overlaps in recent history.
Importantly, repeats act on both sides of the decision: they make a subset of \emph{non-match} trials confusable (lure non-matches)~\citep{szmalec2011control}, and in small grids they can also suppress evidence on true \emph{match} trials (lure matches)~\citep{ni2024computational}, jointly producing block-level collapse toward an ``always non-match'' response policy.

These findings also sharpen how multimodal ``working memory'' should be interpreted in contemporary VLMs.
Humans can flexibly recruit visuospatial attention and rehearsal strategies during \textit{n}-back~\citep{postle2005effects, kane2007working}, whereas many current VLMs are formed by attaching a vision encoder and projection to a pretrained text-only LLM \citep{liu2023visual, qwen2.5-VL}.
Our results are consistent with the view that this visual route may be less robust under load for the control operations that \textit{n}-back is designed to elicit---online updating, lagged temporal binding, and interference control---even when the task-relevant information is matched.
More broadly, they suggest that multimodal evaluation should be computation-sensitive: it is not sufficient to show that a model can accept multimodal inputs; we must test whether it carries out the intended operations of classic paradigms.

We intentionally used simple, tightly controlled prompts and did not introduce prompt-based scaffolding that has been reported to improve apparent \textit{n}-back performance in text-only settings \citep{gong2024working, zhang2024working}.
It remains unclear whether analogous scaffolding would help for vision-grid inputs, and whether any gains would reflect improved temporal binding versus a shift to alternative heuristics.
Beyond prompting, our study varies representational code only between text and vision.
Future work should extend the same controlled \textit{n}-back design to audio in newer audio-capable multimodal models~\citep{xu2025qwen2, hurst2024gpt}, enabling closer alignment with human auditory \textit{n}-back variants and testing whether modality gaps generalize beyond vision~\citep{amon2018auditory}.

Finally, it is important to distinguish \emph{working memory} from long-context management.
Systems that render histories into images for OCR-based compression or externalized state \citep{wei2025deepseek, feng2026agentocr} primarily target storage and retrieval over long horizons, rather than the rapid, interference-sensitive updating probed by \textit{n}-back.
Our results therefore do not challenge the utility of visual compression for long-context management; rather, they caution against assuming that visual-token substitution preserves working-memory-like computations under rapid updating demands.
In short, using images to store \emph{more} context is not the same as using vision to support \emph{online} updating and temporal binding.

Overall, representational code matters in a computation-relevant sense: in matched spatial \textit{n}-back probes, replacing text with vision substantially impairs performance. Diagnostic analyses reveal recency-locked strategies and interference-driven failures invisible to standard accuracy metrics.
This supports a more careful, computation-level interpretation of ``multimodal working memory'' in current VLMs and motivates more explicit tests of the operations that classic WM paradigms are intended to isolate. 

\newpage
\printbibliography

@article{baddeley2020working,
  title={Working memory},
  author={Baddeley, Alan},
  journal={Memory},
  pages={71--111},
  year={2020},
  publisher={Routledge}
}

@incollection{BADDELEY197447,
title = {Working Memory},
editor = {Gordon H. Bower},
series = {Psychology of Learning and Motivation},
publisher = {Academic Press},
volume = {8},
pages = {47-89},
year = {1974},
issn = {0079-7421},
doi = {https://doi.org/10.1016/S0079-7421(08)60452-1},
url = {https://www.sciencedirect.com/science/article/pii/S0079742108604521},
author = {Alan D. Baddeley and Graham Hitch}
}

@article{kane2000working,
  title={Working-memory capacity, proactive interference, and divided attention: limits on long-term memory retrieval.},
  author={Kane, Michael J and Engle, Randall W},
  journal={Journal of Experimental Psychology: Learning, Memory, and Cognition},
  volume={26},
  number={2},
  pages={336},
  year={2000},
  publisher={American Psychological Association}
}

@inproceedings{zhang2024working,
  title={Working memory identifies reasoning limits in language models},
  author={Zhang, Chunhui and Jian, Yiren and Ouyang, Zhongyu and Vosoughi, Soroush},
  booktitle={Proceedings of the 2024 Conference on Empirical Methods in Natural Language Processing},
  pages={16896--16922},
  year={2024}
}

@article{frost2021n,
  title={Is the n-back task a measure of unstructured working memory capacity? Towards understanding its connection to other working memory tasks},
  author={Frost, Adam and Moussaoui, Simar and Kaur, Jagjot and Aziz, Samreen and Fukuda, Keisuke and Niemeier, Matthias},
  journal={Acta psychologica},
  volume={219},
  pages={103398},
  year={2021},
  publisher={Elsevier}
}

@article{kirchner1958age,
  title={Age differences in short-term retention of rapidly changing information.},
  author={Kirchner, Wayne K},
  journal={Journal of experimental psychology},
  volume={55},
  number={4},
  pages={352},
  year={1958},
  publisher={American Psychological Association}
}

@book{hautus2021detection,
  title={Detection theory: A user's guide},
  author={Hautus, Michael J and Macmillan, Neil A and Creelman, C Douglas},
  year={2021},
  publisher={Routledge}
}

@inproceedings{gong2024working,
  title={Working memory capacity of ChatGPT: An empirical study},
  author={Gong, Dongyu and Wan, Xingchen and Wang, Dingmin},
  booktitle={Proceedings of the AAAI conference on artificial intelligence},
  volume={38},
  number={9},
  pages={10048--10056},
  year={2024}
}

@article{qwen2.5,
    title   = {Qwen2.5 Technical Report}, 
    author  = {An Yang and Baosong Yang and Beichen Zhang and Binyuan Hui and Bo Zheng and Bowen Yu and Chengyuan Li and Dayiheng Liu and Fei Huang and Haoran Wei and Huan Lin and Jian Yang and Jianhong Tu and Jianwei Zhang and Jianxin Yang and Jiaxi Yang and Jingren Zhou and Junyang Lin and Kai Dang and Keming Lu and Keqin Bao and Kexin Yang and Le Yu and Mei Li and Mingfeng Xue and Pei Zhang and Qin Zhu and Rui Men and Runji Lin and Tianhao Li and Tingyu Xia and Xingzhang Ren and Xuancheng Ren and Yang Fan and Yang Su and Yichang Zhang and Yu Wan and Yuqiong Liu and Zeyu Cui and Zhenru Zhang and Zihan Qiu},
    journal = {arXiv preprint arXiv:2412.15115},
    year    = {2024}
}

@article{qwen2.5-VL,
  title={Qwen2. 5-vl technical report},
  author={Bai, Shuai and Chen, Keqin and Liu, Xuejing and Wang, Jialin and Ge, Wenbin and Song, Sibo and Dang, Kai and Wang, Peng and Wang, Shijie and Tang, Jun and others},
  journal={arXiv preprint arXiv:2502.13923},
  year={2025}
}

@article{hurst2024gpt,
  title={Gpt-4o system card},
  author={Hurst, Aaron and Lerer, Adam and Goucher, Adam P and Perelman, Adam and Ramesh, Aditya and Clark, Aidan and Ostrow, AJ and Welihinda, Akila and Hayes, Alan and Radford, Alec and others},
  journal={arXiv preprint arXiv:2410.21276},
  year={2024}
}

@article{team2024gemini,
  title={Gemini 1.5: Unlocking multimodal understanding across millions of tokens of context},
  author={Team, Gemini and Georgiev, Petko and Lei, Ving Ian and Burnell, Ryan and Bai, Libin and Gulati, Anmol and Tanzer, Garrett and Vincent, Damien and Pan, Zhufeng and Wang, Shibo and others},
  journal={arXiv preprint arXiv:2403.05530},
  year={2024}
}

@article{wei2025deepseek,
  title={Deepseek-ocr: Contexts optical compression},
  author={Wei, Haoran and Sun, Yaofeng and Li, Yukun},
  journal={arXiv preprint arXiv:2510.18234},
  year={2025}
}

@article{feng2026agentocr,
  title={AgentOCR: Reimagining Agent History via Optical Self-Compression},
  author={Feng, Lang and Yang, Fuchao and Chen, Feng and Cheng, Xin and Xu, Haiyang and Wan, Zhenglin and Yan, Ming and An, Bo},
  journal={arXiv preprint arXiv:2601.04786},
  year={2026}
}

@article{liu2023visual,
  title={Visual instruction tuning},
  author={Liu, Haotian and Li, Chunyuan and Wu, Qingyang and Lee, Yong Jae},
  journal={Advances in neural information processing systems},
  volume={36},
  pages={34892--34916},
  year={2023}
}

@inproceedings{
xing2025visioncentric,
title={Vision-centric Token Compression in Large Language Model},
author={Ling Xing and Alex Jinpeng Wang and Rui Yan and Xiangbo Shu and Jinhui Tang},
booktitle={The Thirty-ninth Annual Conference on Neural Information Processing Systems},
year={2025},
url={https://openreview.net/forum?id=YdggdEL41C}
}

@inproceedings{hu-lewis-2025-language,
    title = "Do Language Models Understand the Cognitive Tasks Given to Them? Investigations with the N-Back Paradigm",
    author = "Hu, Xiaoyang  and
      Lewis, Richard",
    editor = "Che, Wanxiang  and
      Nabende, Joyce  and
      Shutova, Ekaterina  and
      Pilehvar, Mohammad Taher",
    booktitle = "Findings of the Association for Computational Linguistics: ACL 2025",
    month = jul,
    year = "2025",
    address = "Vienna, Austria",
    publisher = "Association for Computational Linguistics",
    url = "https://aclanthology.org/2025.findings-acl.136/",
    doi = "10.18653/v1/2025.findings-acl.136",
    pages = "2665--2677",
    ISBN = "979-8-89176-256-5"
}

@inproceedings{hong-etal-2025-exploring,
    title = "Exploring Working Memory Capacity in {LLM}s: From Stressors to Human-Inspired Strategies",
    author = "Hong, Eunjin  and
      Cho, Sumin  and
      Kim, Juae",
    editor = "Inui, Kentaro  and
      Sakti, Sakriani  and
      Wang, Haofen  and
      Wong, Derek F.  and
      Bhattacharyya, Pushpak  and
      Banerjee, Biplab  and
      Ekbal, Asif  and
      Chakraborty, Tanmoy  and
      Singh, Dhirendra Pratap",
    booktitle = "Proceedings of the 14th International Joint Conference on Natural Language Processing and the 4th Conference of the Asia-Pacific Chapter of the Association for Computational Linguistics",
    month = dec,
    year = "2025",
    address = "Mumbai, India",
    publisher = "The Asian Federation of Natural Language Processing and The Association for Computational Linguistics",
    url = "https://aclanthology.org/2025.ijcnlp-long.93/",
    pages = "1727--1744",
    ISBN = "979-8-89176-298-5",
}

@article{JANSMA2000688,
title = {Specific versus Nonspecific Brain Activity in a Parametric N-Back Task},
journal = {NeuroImage},
volume = {12},
number = {6},
pages = {688-697},
year = {2000},
issn = {1053-8119},
doi = {https://doi.org/10.1006/nimg.2000.0645},
url = {https://www.sciencedirect.com/science/article/pii/S1053811900906451},
author = {Johan Martijn Jansma and Nick F. Ramsey and Richard Coppola and René S. Kahn}
}

@article{owen2005n,
  title={N-back working memory paradigm: A meta-analysis of normative functional neuroimaging studies},
  author={Owen, Adrian M and McMillan, Kathryn M and Laird, Angela R and Bullmore, Ed},
  journal={Human brain mapping},
  volume={25},
  number={1},
  pages={46--59},
  year={2005},
  publisher={Wiley Online Library}
}

@article{mcmillan2007self,
  title={Self-paced working memory: Validation of verbal variations of the n-back paradigm},
  author={McMillan, Kathryn M and Laird, Angela R and Witt, Suzanne T and Meyerand, M Elizabeth},
  journal={Brain research},
  volume={1139},
  pages={133--142},
  year={2007},
  publisher={Elsevier}
}

@article{postle2005effects,
  title={Effects of verbal and nonverbal interference on spatial and object visual working memory},
  author={Postle, Bradley R and D’Esposito, Mark and Corkin, Suzanne},
  journal={Memory \& cognition},
  volume={33},
  number={2},
  pages={203--212},
  year={2005},
  publisher={Springer}
}

@article{jaeggi2008improving,
  title={Improving fluid intelligence with training on working memory},
  author={Jaeggi, Susanne M and Buschkuehl, Martin and Jonides, John and Perrig, Walter J},
  journal={Proceedings of the National Academy of Sciences},
  volume={105},
  number={19},
  pages={6829--6833},
  year={2008},
  publisher={National Academy of Sciences}
}

@article{brown2020language,
  title={Language models are few-shot learners},
  author={Brown, Tom and Mann, Benjamin and Ryder, Nick and Subbiah, Melanie and Kaplan, Jared D and Dhariwal, Prafulla and Neelakantan, Arvind and Shyam, Pranav and Sastry, Girish and Askell, Amanda and others},
  journal={Advances in neural information processing systems},
  volume={33},
  pages={1877--1901},
  year={2020}
}

@article{fawcett2006introduction,
  title={An introduction to ROC analysis},
  author={Fawcett, Tom},
  journal={Pattern recognition letters},
  volume={27},
  number={8},
  pages={861--874},
  year={2006},
  publisher={Elsevier}
}

@article{zhou2025towards,
  title={Towards Long-window Anchoring in Vision-Language Model Distillation},
  author={Zhou, Haoyi and Li, Shuo and Chen, Tianyu and Song, Qi and Gao, Chonghan and Li, Jianxin},
  journal={arXiv preprint arXiv:2512.21576},
  year={2025}
}

@book{marr2010vision,
  title={Vision: A computational investigation into the human representation and processing of visual information},
  author={Marr, David},
  year={2010},
  publisher={MIT press}
}

@article{wang2024survey,
  title={A survey on large language model based autonomous agents},
  author={Wang, Lei and Ma, Chen and Feng, Xueyang and Zhang, Zeyu and Yang, Hao and Zhang, Jingsen and Chen, Zhiyuan and Tang, Jiakai and Chen, Xu and Lin, Yankai and others},
  journal={Frontiers of Computer Science},
  volume={18},
  number={6},
  pages={186345},
  year={2024},
  publisher={Springer}
}

@article{10.1561/0600000110,
author = {Li, Chunyuan and Gan, Zhe and Yang, Zhengyuan and Yang, Jianwei and Li, Linjie and Wang, Lijuan and Gao, Jianfeng},
title = {Multimodal Foundation Models: From Specialists to General-Purpose Assistants},
year = {2024},
issue_date = {May 2024},
publisher = {Now Publishers Inc.},
address = {Hanover, MA, USA},
volume = {16},
number = {1–2},
issn = {1572-2740},
url = {https://doi.org/10.1561/0600000110},
doi = {10.1561/0600000110},
month = may,
pages = {1–214},
numpages = {217}
}

@article{liu2024lost,
  title={Lost in the middle: How language models use long contexts},
  author={Liu, Nelson F and Lin, Kevin and Hewitt, John and Paranjape, Ashwin and Bevilacqua, Michele and Petroni, Fabio and Liang, Percy},
  journal={Transactions of the association for computational linguistics},
  volume={12},
  pages={157--173},
  year={2024}
}

@article{dubey2024llama,
  title={The llama 3 herd of models},
  author={Dubey, Abhimanyu and Jauhri, Abhinav and Pandey, Abhinav and Kadian, Abhishek and Al-Dahle, Ahmad and Letman, Aiesha and Mathur, Akhil and Schelten, Alan and Yang, Amy and Fan, Angela and others},
  journal={arXiv e-prints},
  pages={arXiv--2407},
  year={2024}
}

@article{schmiedek2009interference,
  title={Interference and facilitation in spatial working memory: age-associated differences in lure effects in the n-back paradigm.},
  author={Schmiedek, Florian and Li, Shu-Chen and Lindenberger, Ulman},
  journal={Psychology and aging},
  volume={24},
  number={1},
  pages={203},
  year={2009},
  publisher={American Psychological Association}
}

@article{kane2007working,
  title={Working memory, attention control, and the N-back task: a question of construct validity.},
  author={Kane, Michael J and Conway, Andrew RA and Miura, Timothy K and Colflesh, Gregory JH},
  journal={Journal of Experimental psychology: learning, memory, and cognition},
  volume={33},
  number={3},
  pages={615},
  year={2007},
  publisher={American Psychological Association}
}

@article{jonides2006brain,
  title={Brain mechanisms of proactive interference in working memory},
  author={Jonides, John and Nee, Derek E},
  journal={Neuroscience},
  volume={139},
  number={1},
  pages={181--193},
  year={2006},
  publisher={Elsevier}
}

@article{xu2025qwen2,
  title={Qwen2. 5-omni technical report},
  author={Xu, Jin and Guo, Zhifang and He, Jinzheng and Hu, Hangrui and He, Ting and Bai, Shuai and Chen, Keqin and Wang, Jialin and Fan, Yang and Dang, Kai and others},
  journal={arXiv preprint arXiv:2503.20215},
  year={2025}
}

@inproceedings{amon2018auditory,
  title={Auditory versus visual stimulus effects on cognitive performance during the N-back task},
  author={Amon, Mary Jean and Bertenthal, Bennett I},
  booktitle={Proceedings of the Annual Meeting of the Cognitive Science Society},
  volume={40},
  year={2018}
}

@inproceedings{harbison2012n,
  title={N-back performance: comparing assessment and training performance},
  author={Harbison, J and Atkins, Sharona and Dougherty, Michael R},
  booktitle={Proceedings of the Annual Meeting of the Cognitive Science Society},
  volume={34},
  number={34},
  year={2012}
}

@article{kiani2009representation,
  title={Representation of confidence associated with a decision by neurons in the parietal cortex},
  author={Kiani, Roozbeh and Shadlen, Michael N},
  journal={science},
  volume={324},
  number={5928},
  pages={759--764},
  year={2009},
  publisher={American Association for the Advancement of Science}
}

@article{yonelinas1994receiver,
  title={Receiver-operating characteristics in recognition memory: evidence for a dual-process model.},
  author={Yonelinas, Andrew P},
  journal={Journal of experimental psychology: Learning, memory, and cognition},
  volume={20},
  number={6},
  pages={1341},
  year={1994},
  publisher={American Psychological Association}
}

@article{szmalec2011control,
  title={Control of interference during working memory updating.},
  author={Szmalec, Arnaud and Verbruggen, Frederick and Vandierendonck, Andr{\'e} and Kemps, Eva},
  journal={Journal of Experimental Psychology: Human Perception and Performance},
  volume={37},
  number={1},
  pages={137},
  year={2011},
  publisher={American Psychological Association}
}

@article{ni2024computational,
  title={A computational approach to the N-back task},
  author={Ni, Long and Ma, Wei Ji},
  journal={Scientific reports},
  volume={14},
  number={1},
  pages={30211},
  year={2024},
  publisher={Nature Publishing Group UK London}
}

@inproceedings{o2022measuring,
  title={Measuring and modeling confidence in human causal judgment},
  author={O’Neill, Kevin and Henne, Paul and Pearson, John and DeBrigard, Felipe},
  booktitle={Proceedings of the Annual Meeting of the Cognitive Science Society},
  volume={44},
  number={44},
  year={2022}
}

@article{baddeley2000episodic,
  title={The episodic buffer: a new component of working memory?},
  author={Baddeley, Alan},
  journal={Trends in cognitive sciences},
  volume={4},
  number={11},
  pages={417--423},
  year={2000},
  publisher={Elsevier}
}

@inproceedings{destefano2020influences,
  title={Influences of both prior knowledge and recent historyon visual working memory},
  author={DeStefano, Isabella and Vul, Edward and Brady, Timothy F},
  booktitle={Proceedings of the Annual Meeting of the Cognitive Science Society},
  volume={42},
  year={2020}
}

@inproceedings{brady2010encoding,
  title={Encoding higher-order structure in visual working memory: A probabilistic model},
  author={Brady, Timothy and Tenebaum, Joshua},
  booktitle={Proceedings of the Annual Meeting of the Cognitive Science Society},
  volume={32},
  number={32},
  year={2010}
}

@inproceedings{10.1145/3746027.3755002,
author = {Zhu, Hongyu and Liang, Sichu and Wang, Wenwen and Li, Boheng and Yuan, Tongxin and Li, Fangqi and Wang, Hanyi and Wang, Shi-Lin and Zhang, Zhuosheng},
title = {Revisiting Data Auditing in Large Vision-Language Models},
year = {2025},
isbn = {9798400720352},
publisher = {Association for Computing Machinery},
address = {New York, NY, USA},
url = {https://doi.org/10.1145/3746027.3755002},
doi = {10.1145/3746027.3755002},
booktitle = {Proceedings of the 33rd ACM International Conference on Multimedia},
pages = {11337–11346},
numpages = {10},
location = {Dublin, Ireland},
series = {MM '25}
}

\end{document}